\documentclass[letterpaper, 10 pt, conference]{ieeeconf}  

\IEEEoverridecommandlockouts                              

\usepackage{url}

\usepackage{graphics} 
\usepackage{amsmath}
\usepackage{amssymb}
\usepackage{booktabs}
\usepackage{graphicx}
\usepackage{tabularx}
\usepackage{dblfloatfix}

\title{\LARGE \bf {
Learning Transferable Motor Skills for Geometry-Aware Robotic Surface Tasks
}}

\author{Miroslav David$^{1}$\thanks{Presented at the Workshop on Geometry in the Age of Data-Driven Robotics, ICRA 2026, Vienna, Austria.}, Karla Stepanova$^{1}$, and Robert Babuska$^{1,2}$%
\thanks{$^{1}$ Department of Robotics and Machine Perception, Czech Institute of Informatics, Robotics, and Cybernetics, Czech Technical University in Prague, Czech Republic.}%
\thanks{$^{2}$ Cognitive Robotics, Faculty of 3mE, Delft University of Technology, The Netherlands.}%
\thanks{This paper was supported by European Union's Horizon Europe research and innovation programme under grant agreement number 101214398 (ELLIOT). Views and opinions expressed are those of the author(s) only and do not necessarily reflect those of the European Union or the European Commission. This work was also supported by the Ministry of Education, Youth and Sports of the Czech Republic through the e-INFRA CZ (ID:90254). This research was co-funded by the European Union under the project Robotics and Advanced Industrial Production (reg. no. CZ.02.01.01/00/22\_008/0004590).}}

\begin{document}

\maketitle
\thispagestyle{empty}
\pagestyle{empty}


\begin{figure*}[t!]
  \centering
  \includegraphics[width=0.90\textwidth]{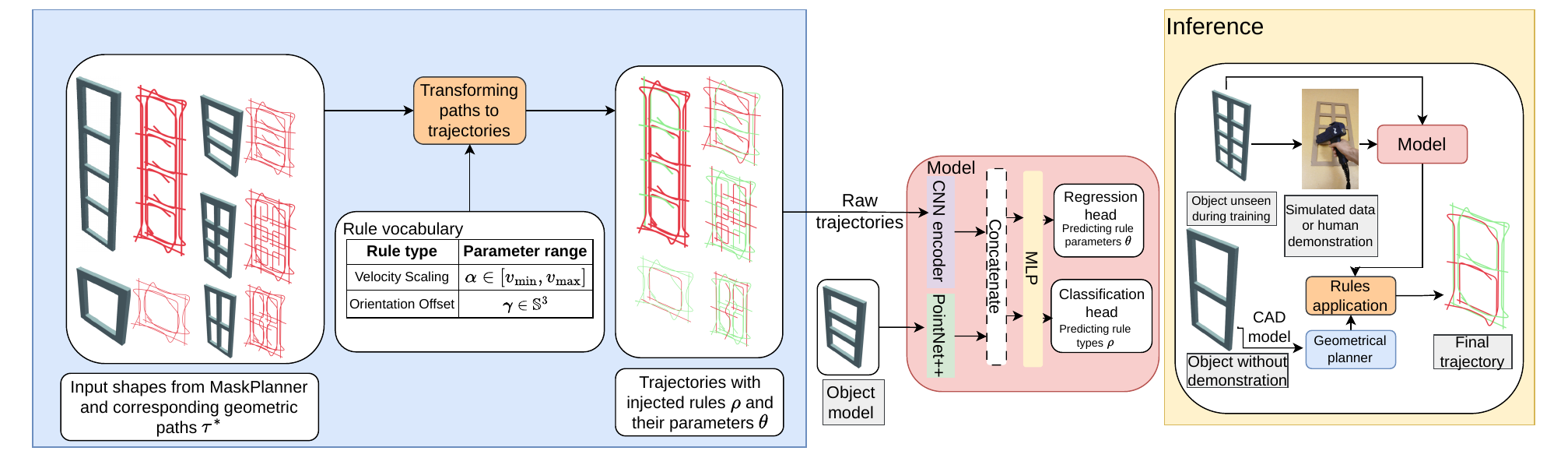}
  \caption{Overview of the proposed rule-based skill injection framework, illustrating the pipeline from synthetic training data generation to skill transfer on novel geometric paths.
}
\label{fig:pipeline}
\end{figure*}

\begin{abstract}
Robotic surface-interaction tasks, such as spray painting or welding, require both accurate geometric planning and precise motion execution. While modern motion planners generate valid geometric paths, they often lack the expert motor patterns observed in human operators. Conversely, learning from demonstration often tightly couples task execution to the specific training geometry, limiting transferability. We propose a modular framework that decouples geometric motion planning from execution-level expertise. Expert behavior is represented as a vocabulary of interpretable, atomic motor rules, such as velocity scaling and orientation offsets, that systematically modify a geometrically planned reference path. We train a multimodal neural network to infer rule parameters jointly from kinematic trajectory data and CAD model geometry. We evaluate our approach through dynamic simulation on L-shaped and window-shaped objects, demonstrating on simulated data that the model successfully extracts velocity and orientation rules across both topologies.
\end{abstract}


\section{INTRODUCTION}

Robotic surface-interaction tasks, such as spray painting or welding, require not only accurate geometric motion planning but also precise execution strategies. While modern motion planners~\cite{tiboniMaskPlannerLearningBasedObjectCentric2025, rabino2025foldpathendtoendobjectcentricmotion, CAD_based_trajectory} can generate collision-free paths that satisfy geometric constraints, the resulting trajectories often lack the nuanced motor patterns observed in human expert behavior. Skilled operators adapt velocity and orientation according to the local geometry of the workpiece, producing execution patterns that are difficult to capture using purely geometric planning methods~\cite{s23249634, Hua2024-ha}. An alternative approach is to directly show the desired robot motions through human demonstrations~\cite{zhao2023learningfinegrainedbimanualmanipulation}. While such an approach preserves expert motor skills, the resulting trajectories are often tightly coupled to the demonstrated geometry and, therefore, difficult to transfer to new workpieces.

In this work, we propose a modular approach for integrating human motor expertise with geometric motion planning. Expert behavior is represented as a set of interpretable motor rules, such as velocity scaling and orientation offsets, that modify the execution of a geometrically planned path, decoupling planning from execution-level expertise. A multimodal neural network combining a 1D convolutional trajectory encoder and a PointNet++~\cite{qi2017pointnet++} CAD encoder is used to infer segment types and rule parameters from trajectory data, and the extracted rules are injected into a physics-based simulation to refine geometrically planned trajectories.

The contributions of this work are: (1) A modular framework that represents expert motor skills as a parametric rule vocabulary and injects them into geometrically planned trajectories. (2) A multimodal neural architecture combining a 1D convolutional trajectory encoder and a PointNet++ CAD encoder to infer rule parameters from demonstration data.

\section{RELATED WORK}
Classical approaches to robotic spray painting primarily focus on geometric coverage and physics-based deposition models~\cite{hepingchenAutomatedRobotTrajectory2002,gleesonGeneratingOptimizedTrajectories2022,lewkeAutomatedTrajectoryPlanning2024}.
%
%
A different line of work addresses skill acquisition through learning from demonstrations and policy learning 
%
~\cite{ratliffMaximumMarginPlanning2006, rossReductionImitationLearning, montgomeryGuidedPolicySearch2016}. These methods 
typically learn a tightly coupled mapping between observed behavior and the task setting in which it is demonstrated. As a result, execution strategies are often tied to the geometry of the demonstrated scenario, limiting transfer to new workpiece shapes or planning contexts.

Movement representation frameworks such as Dynamic Movement Primitives (DMPs) offer a compact way to encode and reproduce motion skills through structured dynamical systems~\cite{ijspeertDynamicalMovementPrimitives2013, saverianoDynamicMovementPrimitives2023}. Probabilistic movement primitives extend this idea by representing distributions over trajectories and enabling adaptation and blending~\cite{paraschos2013probabilistic}. These approaches provide reusable motion representations, but they typically encode behavior at the trajectory level and are less naturally suited to expressing isolated, interpretable execution rules tied to local geometric context.

More recently, data-driven trajectory generation and refinement methods have explored neural approaches to modify or generate robot motion. Residual learning methods augment nominal controllers with learned corrections~\cite{johanninkResidualReinforcementLearning2019, paneReinforcementLearningBased2019}, while transformer-based models have been used to reshape existing trajectories from multimodal inputs such as natural language \cite{9981810}. In robotic spray painting specifically, recent learning-based planners such as PaintNet~\cite{tiboniPaintNetUnstructuredMultiPath2023} and MaskPlanner~\cite{tiboniMaskPlannerLearningBasedObjectCentric2025} predict geometric path structures directly from 3D object representations. These methods demonstrate strong generalization in geometric path generation but often rely on implicit latent representations and do not explicitly separate execution-level expertise from the underlying geometric plan.

\section{PROBLEM FORMULATION}
\label{sec:problem_formulation}
Let $\mathcal{T}$ denote the space of all possible trajectories and $\mathcal{O}$ denote the space of all workpiece geometries. Each object geometry $O \in \mathcal{O}$ is represented as a point cloud $O = \{\mathbf{x}_1, \mathbf{x}_2, \ldots, \mathbf{x}_K\}$, where each point $\mathbf{x}_k \in \mathbb{R}^3$. A set of expert demonstrations is defined as $\mathcal{D} = \{(\tau_1, O_1), \ldots, (\tau_n, O_n) \}$. Each trajectory demonstration $\tau_i = \{(t_j, \mathbf{p}_j, \mathbf{q}_j, v_j)\}_{j=1}^{T_i}$ is a time-indexed sequence consisting of positions $\mathbf{p}_j \in \mathbb{R}^3$, orientations as unit quaternions $\mathbf{q}_j \in \mathbb{S}^3$, and linear speed $v_j \in \mathbb{R}$.

Given $\mathcal{D}$, our goals are to: (1) extract a set of interpretable rules $\mathcal{R} = \{\rho_1, \ldots, \rho_m\}$ and associated parameter spaces $\Theta$ that characterize expert skills relative to the workpiece geometry, (2) learn an inference function $f: \mathcal{T} \times \mathcal{O} \;\rightarrow\; \mathcal{C}^m \times \Theta$ that, for each rule $\rho_k$, predicts a target geometric segment class $c_k \in \mathcal{C} = \{\texttt{straight}, \texttt{corner}, \texttt{none}\}$ and a continuous execution parameter $\theta_k \in \Theta$, and (3) inject the inferred rules into a raw geometric trajectory $\tau^*$ for a new workpiece $O^*$. A rule $\rho_k$ is considered active when $c_k \neq \texttt{none}$. The skill-enhanced trajectory $\hat{\tau}$ is then produced as $\hat{\tau} = (\rho_{n}(\theta_{n}, O^*) \circ \dots \circ \rho_{1}(\theta_{1}, O^*))(\tau^*)$, where only active rules are applied to their corresponding segment type.

\section{METHODOLOGY}

This section details the formal representation of rules and the underlying physics governing the trajectory refinement process. 
An overview of the proposed pipeline is shown in Fig.~\ref{fig:pipeline}, illustrating the artificial data generation process, rule inference training, and skill transfer to unseen geometries.

\subsection{Rule-Based Skill Representation}
\label{sec:rule_definition}
Expert skills are modeled as a structured vocabulary of parametric transformations. The raw geometric path $\tau^*$ is first decomposed into a sequence of spatial segments $\mathcal{S}(\tau^*) = \{s_1, s_2, \ldots, s_L\}$, where each segment $s_i$ is described by a set of geometric and spatial features $c_i$ (e.g., shape type such as straight or corner).

The trajectory profile for each segment can then be systematically modified by applicable expert rules. A general rule $\rho_k \in \mathcal{R}$ consists of an \textit{activation condition} evaluated against the segment features $c_i$, and a \textit{dynamic override} parameterized by $\theta_k$. These overrides apply systematic modifications to velocity or orientation for each matching segment type.
\subsection{Trajectory Refinement via Dynamic Simulation}
To ensure physically continuous dynamics and avoid instantaneous velocity jumps, a pure pursuit controller tracks a lookahead pose along the reference path, applying independent proportional forces and torques to position and orientation errors. The resulting wrench is numerically integrated via fourth-order Runge-Kutta (RK4), modeling the end-effector as a rigid body with mass, inertia, and viscous damping.
\subsection{Data Generation and Preprocessing}
Training data are generated using custom L-shaped topologies alongside MaskPlanner window geometries \cite{tiboniMaskPlannerLearningBasedObjectCentric2025} to evaluate generalization across varying surface orientations and structural crossings. Raw geometric paths $(x, y, z, \text{Euler angles}, part ID)$ are segmented by part ID, and the orientations are converted to unit quaternions for kinematic simulation. These static paths are then processed through our simulation to yield dynamically rich, time-indexed trajectories $\tau = \{(t_j, \mathbf{p}_j, \mathbf{q}_j, v_j)\}_{j=1}^T$.
\subsection{Instantiated Rule Vocabulary}
\label{sec:specific_rules}
We define a core rule vocabulary $\mathcal{R}$ consisting of parametric transformations that encode foundational expert painting skills. Rather than modeling full trajectories, these rules conditionally modify the local execution profile based on the current geometric segment type (e.g., straight paths or corners). Specifically, velocity scaling ($\rho_{vel}$) applies a continuous scaling factor to the target linear speed, while orientation offsets ($\rho_{ori}$) introduce systematic tilts (represented as a quaternion offset $\boldsymbol{\gamma} \in \mathbb{S}^3$) relative to the surface normal to mimic an expert's tool tilt.
\subsection{Multimodal Rule Inference}
To infer the applied rules from demonstrations, we employ a multimodal neural architecture. The model takes a kinematic trajectory and the underlying target CAD model as inputs. To ensure stable neural network training, the trajectory orientations are mapped to a continuous 6D rotation representation \cite{zhou2020continuityrotationrepresentationsneural}. The resulting 10-dimensional trajectory features, comprising a time delta (1D), a 3D linear velocity vector, and the 6D rotation representation, are processed by a 1D convolutional encoder comprising three stacked Conv1d layers followed by masked average pooling over valid timesteps, producing a compact trajectory embedding. The CAD model is sampled into a point cloud and encoded by a PointNet++ encoder, yielding a global geometric feature vector. The two modality embeddings are combined via concatenation and passed through a shared MLP, enabling the network to jointly reason over local execution dynamics and the broader geometric context. Although the current rule vocabulary does not directly condition on workpiece features, the CAD encoder is included to support future proximity-based rules that require spatial context from the object structure. Finally, the network utilises two output heads: a regression head that predicts the numerical parameter of the applied rule (e.g., the specific speed multiplier or tilt angle), and a classification head that predicts the geometric segment type to which the rule was applied.
\subsection{Transfer to Unseen Geometries}
\label{sec:transfer}
Given the inferred rules and their target segment types, skill transfer is executed on a geometrically planned path $\tau^*$ for a novel workpiece. The path is segmented automatically: a sliding window of fixed width is swept along the path, and SVD is applied to each windowed point set to measure local curvature; points whose residual distance from the dominant principal axis falls below a threshold are labelled \texttt{straight}, and the remainder \texttt{corner}. No manual annotation is required, making the procedure applicable to any new geometry. The inferred rules are then injected into their corresponding segment types during simulation, yielding a refined trajectory $\hat{\tau}$ that preserves the required geometric structure while incorporating the expert's motor strategies.

\section{EXPERIMENTAL RESULTS}
We assess rule classification and parameter regression across simulated datasets. For simpler L-shaped geometries, 10{,}000 synthetic trajectories were generated and split 8{,}000\,/\,1{,}000\,/\,1{,}000 (train\,/\,val\,/\,test). For more complex window geometries, the total of 1{,}000 trajectories were split 800\,/\,100\,/\,100 (train/val/test).

\begin{figure}[htbp]
\centering
\includegraphics[width=0.5\textwidth]{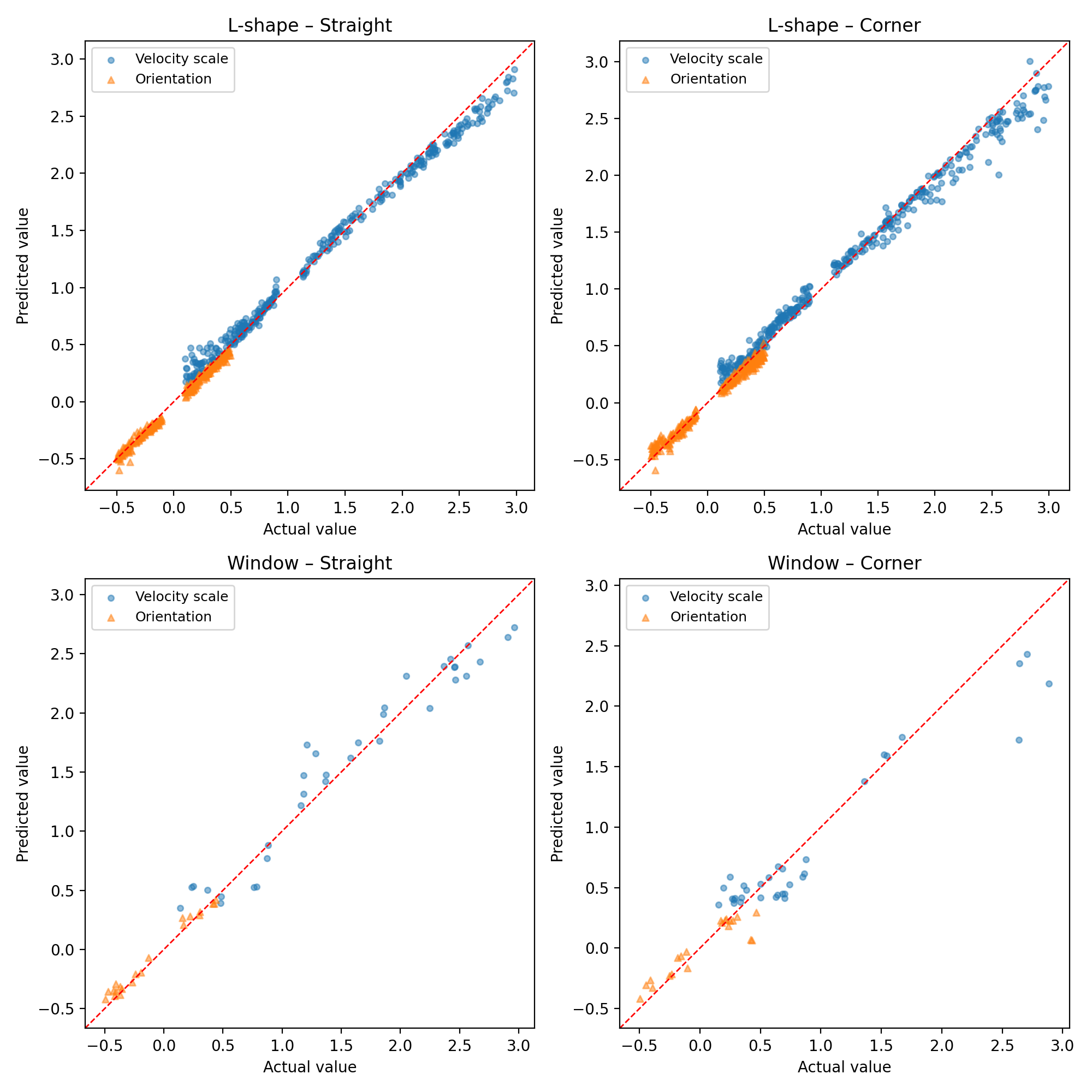}
\caption{Predicted vs.\ actual parameter values on the test set. Each subplot shows one geometry type: L-shape straight (top-left), L-shape corner (top-right), window straight (bottom-left), window corner (bottom-right). Blue circles: velocity scaling factor; orange triangles: orientation offset (rad). The red dashed line indicates perfect prediction.}
\label{fig:regression_scatter}
\vspace{-2em}
\end{figure}

\textbf{Simulated Rule Classification:} On the simpler L-shape geometries, the multimodal network achieves a combined F1-score of 0.999 across three classes (straight, corner, none) (see Fig.~\ref{fig:regression_scatter}, top). On the more complex window geometries (see Fig.~\ref{fig:regression_scatter}, bottom), the model achieves F1-scores of 1.000 for velocity scaling and 0.984 for orientation rules, showing that the model generalises well to more complex geometries.

\textbf{Simulated Parameter Regression:} Continuous parameter extraction is evaluated via Mean Absolute Error (MAE) between predicted and actual values. On L-shapes, the network achieves MAEs of 0.020\,rad (orientation offset) and 0.056 (velocity scaling factor, dimensionless, range 0.1–3.0). On window topologies, the model maintains accurate orientation estimation (MAE of 0.038\,rad) while velocity scaling shows a moderate error of 0.131 (dimensionless), reflecting the shorter and more rapidly changing segments characteristic of window frame geometries. This suggests that finer-grained segmentation or segment-length normalisation may be needed to handle geometries with dense curvature changes. Overall, the simulated results confirm that the architecture successfully disentangles dynamic overrides from the geometric trajectory.

\section{DISCUSSION AND FUTURE WORK}
We demonstrated that execution strategies can be successfully decoupled from workpiece geometry using interpretable, atomic rules in both L-shaped and window topologies. The current rule vocabulary operates purely on segment type and does not yet exploit workpiece geometry at inference time; rules conditioned on runtime proximity to CAD features such as edges or crossings would require the network to perform structured spatial reasoning, which motivates the inclusion of the CAD encoder in anticipation of this extension. Bridging the sim-to-real gap remains the critical next step: physical teaching interfaces currently suffer from low sampling rates, limiting capture of high-frequency motor adjustments. Future work will address this bottleneck, expand the rule vocabulary to include proximity-based rules conditioned on structural CAD features such as crossings or edges, and evaluate zero-shot transfer to held-out topologies such as U-shaped and T-shaped geometries.

\bibliographystyle{IEEEtran}

\bibliography{bibliography_icra2026}  

\end{document}